\documentclass[a4paper,fleqn]{cas-dc}
\makeatletter
\let\showhyphens\@undefined
\makeatother
\usepackage{amssymb}   
\usepackage{amsmath}
\usepackage[numbers]{natbib}
\usepackage{textcomp}
\usepackage{nicematrix}
\usepackage{makecell}
\usepackage{caption}
\usepackage{array}
\usepackage{tabularx}
\usepackage[final]{microtype}
\usepackage{booktabs}
\usepackage{multirow}  
\usepackage{siunitx}   
\usepackage{xcolor}
\usepackage{graphicx}
\usepackage{caption}
\usepackage{float}
\def\tsc#1{\csdef{#1}{\textsc{\lowercase{#1}}\xspace}}
\tsc{WGM}
\tsc{QE}

\begin{document}
\let\WriteBookmarks\relax
\renewcommand{\topfraction}{0.9}
\renewcommand{\bottomfraction}{0.8}
\renewcommand{\textfraction}{0.1}
\renewcommand{\floatpagefraction}{0.7}
\let\printorcid\relax 

\shorttitle{}
\shortauthors{Yue Cao et al.}

\title[mode = title]{Enhancing Feature Fusion of U-like Networks with Dynamic Skip Connections}  

\author[1]{Yue Cao}
\ead{caoyue1@stu.scu.edu.cn}
\author[1]{Quansong He}
\author[1]{Kaishen Wang}
\author[1]{Jianlong Xiong}
\author[1]{Zhang Yi}
\author[1]{Tao He}
\ead{tao_he@scu.edu.cn}
\cormark[1]

\address[1]{College of Computer Science, Sichuan University, Chengdu, 610065, China} 
\cortext[1]{Corresponding author}  

\begin{abstract}
U-like networks have become fundamental frameworks in medical image segmentation through skip connections that bridge high-level semantics and low-level spatial details. Despite their success, conventional skip connections exhibit two key limitations: inter-feature constraints and intra-feature constraints. The inter-feature constraint refers to the static nature of feature fusion in traditional skip connections, where information is transmitted along fixed pathways regardless of feature content. The intra-feature constraint arises from the insufficient modeling of multi-scale feature interactions, thereby hindering the effective aggregation of global contextual information. To overcome these limitations, we propose a novel Dynamic Skip Connection (DSC) block that fundamentally enhances cross-layer connectivity through adaptive mechanisms. The DSC block integrates two complementary components: (1) Test-Time Training (TTT) module: This module addresses the inter-feature constraint by enabling dynamic adaptation of hidden representations during inference, facilitating content-aware feature refinement. (2) Dynamic Multi-Scale Kernel (DMSK) module: To mitigate the intra-feature constraint, this module adaptively selects kernel sizes based on global contextual cues, enhancing the network’s capacity for multi-scale feature integration. The DSC block is architecture-agnostic and can be seamlessly incorporated into existing U-like network structures. Extensive experiments demonstrate the plug-and-play effectiveness of the proposed DSC block across CNN-based, Transformer-based, hybrid CNN-Transformer, and Mamba-based U-like networks. The code is available at \underline{https://github.com/BlackJack-Cao/U-like-Networks-with-DSC}.
\end{abstract}




\maketitle

\section{Introduction}
Medical image segmentation has emerged as a fundamental task in contemporary healthcare applications. It serves as an essential component in clinical diagnosis, preoperative planning, and disease monitoring. The proliferation of medical imaging modalities, including computed tomography (CT), magnetic resonance imaging (MRI), ultrasound, and dermoscopy, has generated unprecedented volumes of high-resolution medical data requiring automated analysis. Traditional manual segmentation approaches, while accurate, are prohibitively time-consuming and subject to inter-observer variability, particularly when dealing with complex anatomical structures or subtle pathological changes. Within this domain, the U-Net architecture~\cite{2015U} has garnered significant attention due to its sophisticated encoder-decoder framework augmented by skip connections, which integrate low-level spatial information from the encoder with high-level semantic representations from the decoder counterpart. 

Despite the success of U-Net, we identify two fundamental constraints that limit the performance of existing architectures. Recent advances have attempted to address the static nature of skip connections. For instance, Attention U-Net~\cite{attentionunet} introduces attention gates to selectively weight features transmitted through skip pathways. However, the attention coefficients in these approaches are computed from fixed feature representations learned during training and remain static during inference, limiting their adaptability to diverse input characteristics. Similarly, while methods like UNet++~\cite{zhou2018unetplusplus} and UNet3+~\cite{2020arXiv200408790H} propose architectural modifications with nested or full-scale skip connections, FuseUNet~\cite{he2025fuseunetmultiscalefeaturefusion} treated skip connections as discrete nodes in an initial value problem framework, they fundamentally rely on predetermined fusion strategies that cannot dynamically adjust based on input-specific content. We define the inter-feature constraints as this fundamental limitation: skip connections maintain static transmission pathways that cannot dynamically modulate based on input-specific characteristics. In medical imaging contexts where anatomical structures and pathological lesions exhibit substantial heterogeneity~\cite{challenges1,challenges2,luo2024stable}, such inflexible connection frameworks cannot adaptively optimize feature propagation.

Beyond the static transmission mechanism, another critical limitation exists within individual skip pathways. This limitation stems from insufficient modeling of adaptive multi-scale feature interactions. Convolutional operations in the main processing streams have incorporated multi-scale mechanisms. For example, SKNet~\cite{SKNet} demonstrates the effectiveness of dynamic kernel selection through attention guided solely by local statistics from its own parallel branches. However, traditional skip connections in U-like networks still rely on fixed kernel sizes for feature integration and lack comparable input-specific global-guided dynamic multi-scale adaptation. Consequently, these fixed-size kernels cannot effectively capture features at different scales~\cite{10.3389/fgene.2021.639930,efc}, which is particularly problematic given the substantial variation in organ sizes and shapes across medical images~\cite{challenges1}. We define this as the intra-feature constraint, which arises from the absence of adaptive multi-scale processing within skip pathways themselves.

To simultaneously address these dual constraints, we propose a Dynamic Skip Connection (DSC) block that integrates Test-Time Training (TTT) module and Dynamic Multi-Scale Kernel (DMSK) module into U-Net architectures. Our contributions are summarized as follows:
\begin{itemize}
\item The DSC block serves as a versatile plug-and-play module demonstrating seamless integration across diverse U-like architectures spanning CNN-based, Transformer-based, hybrid CNN-Transformer, and Mamba-based networks.
\item Different from existing methods that integrate TTT within encoder or decoder components, we pioneer the application of TTT to skip connections. We transform static pathways into adaptive mechanisms that dynamically modulate internal weights based on input-specific characteristics during inference.
\item The DMSK module introduces adaptive kernel size selection based on global contextual cues. It effectively mitigates intra-feature constraints through dynamic multi-scale feature extraction. Meanwhile, it enhances feature representation via integrated channel-wise attention and spatial information aggregation.
\end{itemize}

\section{Related Work}

\subsection{The Development of Skip-connection}
\subsubsection{Implicit Skip-Connection Enhancement Methods}
Implicit skip-connection enhancement methods improve the overall effectiveness of skip connections by enhancing the quality of features generated in the encoder. These enhanced features are subsequently transmitted through skip pathways. TransUNet~\cite{transunet} enhanced the encoder with Transformer-based feature processing. The improved global feature representations from the hybrid CNN-Transformer encoder provided richer contextual information for subsequent skip connection transmission. SwinUNet~\cite{SwinUnet} improved feature representation quality by employing hierarchical shift-window attention mechanisms in the encoder which generated more discriminative multi-scale features that were subsequently transmitted through skip pathways to the decoder. HiFormer~\cite{DBLP:conf/wacv/HeidariKKAACM23} combined CNN and Transformer pathways in the encoder with cross-scale feature fusion mechanisms. The cross-scale feature fusion approach improved the quality of features available for skip connection transmission and enhanced the overall information flow between encoder and decoder components.

\subsubsection{Explicit Skip-Connection Modification Methods}
Explicit skip-connection modification methods focus on enhancing feature integration by restructuring or augmenting the skip pathways themselves. GridNet~\cite{2017Residual} introduced a generalized encoder-decoder framework that utilized grid-structured connections between feature maps to enhance feature aggregation at multiple scales. Similarly, U-Net++~\cite{zhou2018unetplusplus} incorporated nested, dense skip pathways that enabled multi-scale feature aggregation through densely connected convolutional blocks. UNet3+~\cite{2020arXiv200408790H} employed full-scale skip connections that explicitly aggregated features from all encoder levels at each stage of the decoder. FuseUNet~\cite{he2025fuseunetmultiscalefeaturefusion} conceptualized skip connections as discrete nodes within an initial value problem framework. 
While implicit methods effectively enhance feature quality and explicit methods improve connection patterns, both implicit and explicit approaches fundamentally rely on static architectures. These architectures have fixed transmission pathways and predetermined kernel sizes that cannot dynamically adapt to input-specific characteristics during inference. This limitation restricts their optimization potential for diverse medical imaging scenarios.
\subsection{Test-Time Training}
Recent advances in deep learning have introduced TTT~\cite{ttt2,2024Learning} as an innovative method characterized by its linear computational complexity and adaptive dynamic parameters during inference. Unlike traditional methods with fixed parameters, TTT employs a self-supervised learning framework that allows continuous optimization based on test data. This enhances the model's ability to adapt to distribution shifts.
Current applications of TTT in medical imaging have demonstrated its versatility. Zhou et al.~\cite{zhou2024tttunetenhancingunettesttime} proposed TTT-UNet, which integrates TTT layers into the encoder of the traditional U-Net architecture. Xu~\cite{DBLP:journals/corr/abs-2410-02523} introduced Med-TTT, a visual backbone network that incorporates TTT layers to enable effective modeling of long-range dependencies with linear computational complexity.
Beyond medical imaging, TTT has been successfully applied to sequential data processing. Yang, Wang, and Ge~\cite{DBLP:journals/corr/abs-2409-19142} developed TTT4Rec for sequential recommendation tasks, where TTT layers facilitate real-time adaptation to dynamic user behaviors. In bioinformatics,  Meng et al.~\cite{DBLP:journals/corr/abs-2410-13257} proposed scFusionTTT for single-cell multi-omics fusion. This approach leverages TTT-based masked autoencoders to handle sequential relationships between genes and proteins.

\begin{figure*}[t]
  \centering
  \includegraphics[width=1.0\linewidth]{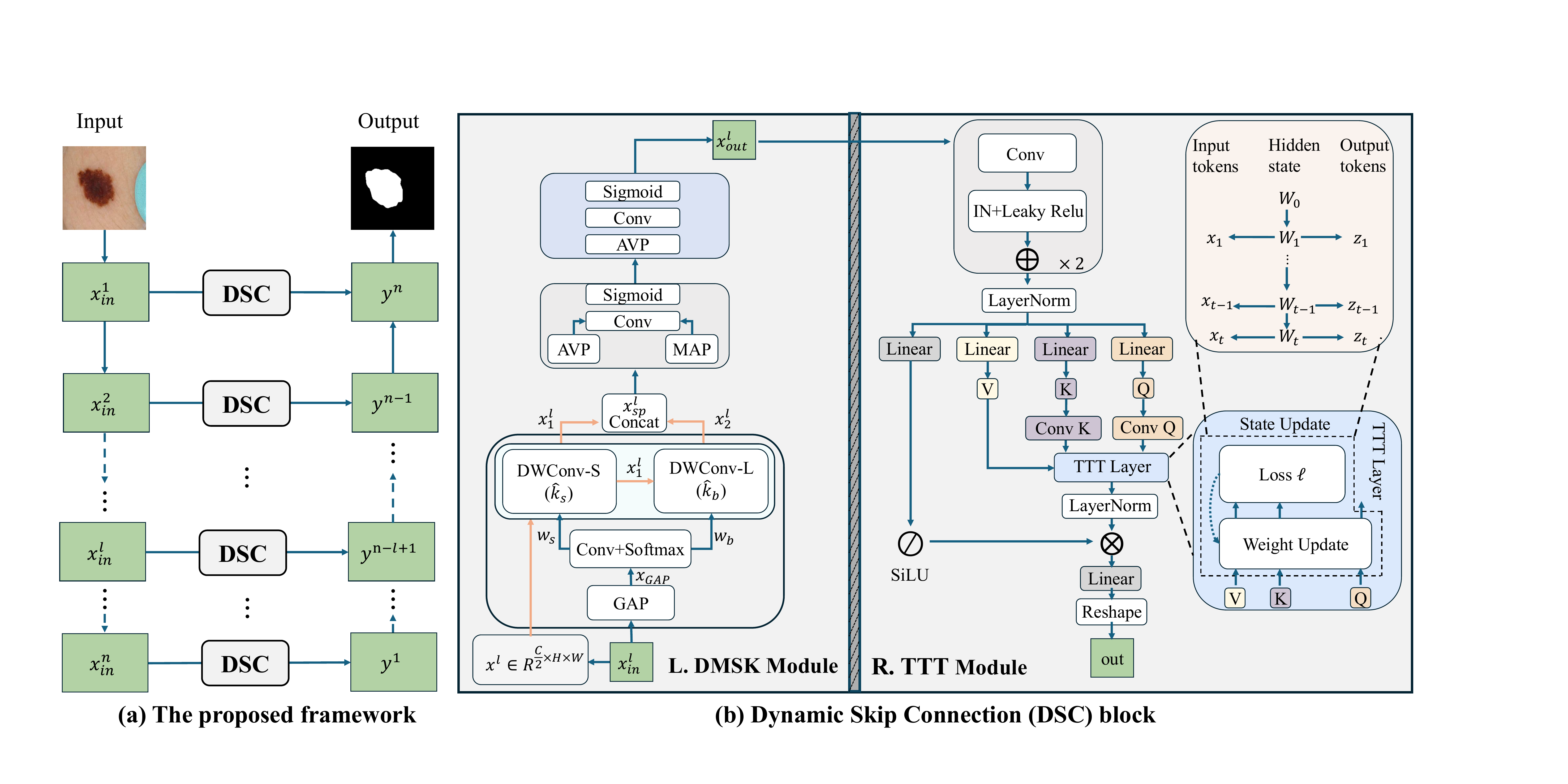}
  \caption{
  (a) Proposed framework with Dynamic Skip Connection (DSC) blocks. (b) DSC block architecture: DMSK module (left) performs adaptive multi-scale kernel selection based on global context; TTT module (right) enables dynamic weight adjustment during inference. Input features ($x^1_{in}, x^2_{in}, \dots, x^n_{in}$) from encoder layers are processed through DMSK and TTT for adaptive feature propagation to the decoder.}
  \label{fig1}
\end{figure*}

\section{Motivation}
Medical image segmentation presents unique challenges that motivate test-time adaptation mechanisms. Unlike natural images where object appearances remain relatively consistent within categories, medical images demonstrate substantial inter-sample heterogeneity even within the same anatomical region and imaging modality. Patient-specific anatomy, pathological variations, imaging protocol differences, and acquisition artifacts create scenarios where each test sample presents characteristics that may significantly deviate from the training distribution. For instance, organ sizes vary dramatically across patients, tissue contrast differs due to disease states, and anatomical boundaries exhibit high shape variability~\cite{challenges1,challenges2}. A critical question then arises: why is test-time adaptation necessary when simpler attention mechanisms could provide feature weighting? The fundamental distinction lies in adaptivity. Attention mechanisms such as Attention U-Net~\cite{attentionunet} compute attention coefficients from static parameters learned during training. These predetermined transformations cannot adjust to novel anatomical configurations or imaging characteristics specific to individual test samples. In contrast, TTT enables genuine sample-specific adaptation by treating each input as a unique learning problem. This per-sample adaptivity necessitates careful consideration of where to integrate such mechanisms within network architectures.

While TTT has demonstrated remarkable capabilities in enabling dynamic parameter adaptation during inference, determining the optimal integration point within U-like architectures presents critical design decisions with distinct theoretical and practical implications. Specifically, the choice between encoder components, decoder components, or skip connection pathways significantly influences the model's adaptive capacity and overall performance. Skip connections in encoder-decoder architectures serve a fundamentally different role compared to encoder or decoder layers. Encoder layers perform progressive feature abstraction within their processing streams, while decoder layers execute semantic-to-spatial reconstruction within their pathways. Skip connections serve as direct feature transmission pathways that preserve spatial details from encoder layers to corresponding decoder layers. However, this direct transmission approach in traditional U-Net architectures is inherently static and cannot adapt to input-specific characteristics. The heterogeneous nature of medical images, characterized by varying anatomical structures and pathological conditions, requires adaptive feature processing at these transmission points to optimize encoder-decoder feature integration. This creates unique optimization opportunities that make skip connections particularly suitable locations for adaptive mechanisms like TTT integration. 

The advantages of applying TTT to skip connection pathways rather than encoder or decoder components are twofold. First, skip connections process encoder features before they are transmitted to decoder layers for integration. TTT can learn input-specific transformations of these encoder features, optimizing their compatibility with decoder processing based on the varying anatomical structures and pathological conditions in medical images. Second, applying TTT to skip connections allows targeted feature adaptation without modifying the encoder and decoder architectures, providing a modular enhancement approach.

Recent works like TTT-UNet~\cite{zhou2024tttunetenhancingunettesttime} have demonstrated TTT effectiveness when applied to encoder components, validating the general utility of TTT in medical image segmentation. However, our analysis suggests that skip connection pathways represent optimal locations for TTT integration due to their unique role in adaptive feature transmission and their computational efficiency advantages.

Dynamic multi-scale kernel approaches excel in adaptively capturing features across different spatial scales, making them particularly valuable for medical image analysis. While prior works such as SKNet~\cite{SKNet} have introduced dynamic kernel selection via attention guided by local branch statistics in the main backbone, MSU-Net~\cite{MSUNet} has employed static multi-branch heterogeneous convolutions with fixed fusion weights primarily in the encoder, and D-Net~\cite{D-Net} has used multiple large kernels with dynamic channel and spatial selection based on global context, our DMSK module distinguishes itself by applying input-specific global-guided dynamic kernel weighting specifically within skip connection pathways for cross-layer feature transmission. This targeted design provides fine-grained, content-aware multi-scale aggregation that seamlessly complements our TTT-enhanced skip connections, enabling adaptive utilization of multi-scale semantic information in both local and global contexts to meet the critical need for input-specific feature adaptation in medical image segmentation.

\section{Method}

\subsection{Framework Structure}
As illustrated in Fig. \ref{fig1}(a), our proposed framework follows an encoder-decoder architecture with enhanced skip connections for medical image segmentation. Let $X\in \mathbb{R}^{C\times H\times W}$ denote the input feature, where $C$, $H$, $W$ represent the number of channels, height, and width, respectively. 

The encoder $\mathcal{E}$ consists of sequential downsampling operations, extracting hierarchical features at multiple resolution levels:
\begin{eqnarray}\{x^1_{in},x^2_{in},...,x^l_{in},...,x^n_{in}\}=\mathcal{E}(X),\end{eqnarray}
where $x^l_{in}$ represents the feature maps at the $l$-th level of the encoder.
Correspondingly, the decoder $\mathcal{D}$ performs progressive upsampling operations to restore the spatial dimensions:
\begin{equation}\{y^1,\ldots,y^{n-l+1},\ldots,y^n\}=\mathcal{D}(\{x^n_{in},\ldots,x^l_{in},\ldots,x^1_{in}\}),\end{equation}
where $y^{n-l+1}$ represents the feature maps at the $l$-th level of the decoder.

We propose the DSC block as the core innovation of our framework. Unlike traditional U-Net architectures that directly concatenate encoder features $x^l_{in}$ with corresponding decoder features $y^{n-l+1}$, our enhanced architecture transforms this pathway by introducing the DSC block to enable adaptive feature processing. The DSC block is designed to simultaneously tackle both inter-feature and intra-feature constraints through dynamic adaptation mechanisms.
As depicted in Fig. \ref{fig1}(b), each DSC block comprises two complementary modules: the DMSK module (Fig. \ref{fig1}(b.L)) and the TTT module (Fig. \ref{fig1}(b.R)). The DMSK module addresses intra-feature constraints by adaptively capturing multi-scale spatial dependencies through dynamic kernel selection, while the TTT module tackles inter-feature constraints by enabling self-supervised adaptation during inference.
The integration of these modules transforms the traditional skip connection pathway as follows:
\begin{equation}y^{n-l+1}=\mathcal{F}_{\mathrm{decode}}(y^{n-l},\mathcal{F}_{\mathrm{TTT}}(\mathcal{F}_{\mathrm{DMSK}}(x^l_{in}))).\end{equation}
When an input feature map $x^l_{in}$ from the encoder enters the DSC block, it first passes through the DMSK module $\mathcal{F}_{\mathrm{DMSK}}$, which processes encoder features to generate multi-scale adaptive representations capturing both local details and global context. These refined features are subsequently processed by the TTT module $\mathcal{F}_{\mathrm{TTT}}$, which implements self-supervised learning during inference to further optimize feature representations based on current input characteristics. The processed features are then integrated into the corresponding decoder level through the decoding function $\mathcal{F}_{\mathrm{decode}}$.
The decoder generates the final segmentation output at full resolution while preserving both fine-grained spatial details and high-level semantic information.

\subsection{Dynamic Multi-Scale Kernel Module}
To address intra-feature constraints, we propose the DMSK module which performs dynamic kernel selection to adaptively capture both local and global spatial dependencies based on input feature characteristics.

To effectively model multi-scale contextual relationships, two sets of kernels are defined: small-scale kernels and large-scale kernels. The former are capable of extracting fine-grained local details while the latter are able to capture long-range dependencies. Each kernel is implemented as a depthwise separable convolution to reduce computational complexity. Furthermore, dilated convolution is applied to expand the receptive field without increasing the number of parameters.
To ensure adaptive selection of the most relevant kernel for each input. Specifically, for the input feature $x_{in}^{l}$, a Global Average Pooling (GAP) operation is first performed to capture global channel-wise statistics:
\begin{equation}x^l_{\mathrm{GAP}}=\mathrm{GAP}(x^{l}_{in}),\end{equation}
the weights for small-scale kernel selection $w_s$ are computed as:
\begin{equation}w_{s}=\mathrm{Softmax}(\mathrm{Conv}_{s}^{l}(x^l_{\mathrm{GAP}})),\end{equation}
where $\mathrm{Conv}_s^{l}$ is a $1\times1$ convolution layer that maps the input channel dimension to the number of small-scale kernels ($|L_s|$). Similarly, the weights for large-scale kernel selection $w_b$ are obtained using a separate convolutional transformation:
\begin{equation}w_b=\mathrm{Softmax}(\mathrm{Conv}_b^{l}(x^l_{\mathrm{GAP}})),\end{equation}
where $\mathrm{Conv}_b^{l}$ is a different $1\times1$ convolution layer mapping the input channels to the number of large-scale kernels ($|L_b|$). 
Consequently, $w_s$ and $w_b$ are indicative of divergent selection probabilities,  thereby enabling the model to autonomously determine the most pertinent small-scale and large-scale kernels for each individual input instance.

Subsequently, a selection method is applied to choose the kernel index. To ensure this discrete selection process is differentiable during training, we employ the Straight-Through Estimator (STE)~\cite{STE}.
\begin{equation}\hat{k}_s=\mathrm{STE}_{\mathrm{argmax}}(w_s),\quad\hat{k}_b=\mathrm{STE}_{\mathrm{argmax}}(w_b),\end{equation}
where $\hat{k}_s$ and $\hat{k}_b$ represent the indices of the selected small-scale and large-scale kernels, respectively. 
Finally, the most relevant small-scale and large-scale convolution kernels are obtained.

Once the most relevant small-scale and large-scale kernels are identified, these kernels are then cascaded with progressively increasing kernel sizes and dilation rates. This design is characterized by two distinct advantages. First, the contextual information is recursively aggregated within the receptive field, thereby enabling the effective receptive field size to gradually expand~\cite{2017Understanding}. Second, features extracted from more extensive and receptive fields contribute more significantly to the output, thereby enabling DMSK to capture more precise and information-rich features. The methodology involves the projection of input features to lower dimensions, followed by the sequential application of kernels that are dynamically selected:
\begin{equation}
\begin{cases}
x^{l}=\mathrm{Project}(x_{in}^l),\\
x_1^l=\mathrm{DWConv}_{(\hat{k}_s)}(x^l),\\
x_2^l=\mathrm{DWConv}_{(\hat{k}_b)}(x_1^l).
\end{cases}
\end{equation}

The resulting features are then concatenated and refined through a spatial attention module that uses both Average Pooling (AVP) and Maximum Pooling (MAP) operations to capture global spatial relationships:
\begin{equation}
\begin{cases}
x_{sp}^l=\mathrm{Concat}([x_1^l;x_2^l]),\\
w_{a}=\mathrm{AVP}(x^l_{sp}),\\
w_{m}=\mathrm{MAP}(x^l_{sp}),\\

[w_{1};w_{2}]=\mathrm{Sigmoid}(\mathrm{Conv}([w_{a};w_{m}])),
\end{cases}
\end{equation}
where both $w_1$ and $w_2$ facilitate the selection of adaptive features from multiple kernel branches. The calibration of feature fusion is achieved through the implementation of element-wise operations:
\begin{equation}x_{ch}^l=(w_1\otimes x_{sp}^l)\oplus(w_2\otimes x_{sp}^l),\end{equation}
where the spatial attention is followed by channel attention that emphasizes feature significance across channels:
\begin{equation}
\begin{cases}
w_{ch} = \mathrm{Sigmoid}(\mathrm{Conv}(\mathrm{AVP}(x^l_{ch})),\\
x_{out}^l=w_{ch}\otimes x_{ch}^l+x_{in}^l,
\end{cases}
\end{equation}
where the channel-wise and spatial attention design complements the dynamic kernel selection by enabling the model to focus on the most informative regions both spatially and channel-wise, significantly enhancing feature representation.
\begin{table*}[h!]
\caption{Dataset information, "\# Targets" specifies distinct segmentation classes in each dataset.}
  \centering
    \setlength{\tabcolsep}{10pt}
    \begin{NiceTabular}{lcccc}
    \toprule
    \textbf{Dataset} & \textbf{Dimension} & \textbf{\# Training Image} & \textbf{\# Testing Image} & \textbf{\# Targets} \\
    \midrule
    ISIC 2017 & 2D    & 1505  & 645   & 2 \\
    Abdomen CT & 3D    & 50 (4794 slices) & 50 (10894 slices) & 13 \\
    Abdomen MRI & 3D    & 60 (5615 slices) & 50 (3357 slices) & 13 \\
    Endoscopy Images & 2D    & 1800  & 1200  & 7 \\
    Microscopy Images & 2D    & 1000  & 101   & 2 \\
    \bottomrule
    \end{NiceTabular}
\label{tab:dataset information}%
\end{table*}%
\subsection{TTT Module}
To mitigate inter-feature constraints, we propose the TTT module which dynamically replaces static skip connections with adaptive feature propagation during inference. It enables real-time adjustment of feature pathways based on input-specific characteristics. As shown in Fig. \ref{fig1}(b.R), the TTT module is essential for facilitating model adaptability during the testing phase. Initially, output features $x^l_{out}$ from the DMSK module are processed through two sequential residual blocks~\cite{2016Deep}. Each block consists of a convolutional layer followed by instance normalization~\cite{2016Instance} and leaky ReLU activation~\cite{Maas0Rectifier}. Subsequently, layer normalization~\cite{LayerNormal} is applied to these features, which are then flattened in preparation for linear transformations. Processing these features through three distinct linear transformation branches yields feature sets designated as $V$, $K$, and $Q$ while a fourth branch implements linear transformation with SiLU activation function~\cite{hendrycks2023gaussianerrorlinearunits} to enrich feature representations. The $V$, $K$, and $Q$ outputs are then transferred to the TTT layer where hidden state updating takes place.

In the TTT layer, the hidden state $h_t$ at time $t$ is parameterized as a trainable model $f$ with dynamically adjustable weights $W_t$. These weights are optimized through gradient descent based on the current input $x_t$:
 \begin{equation}W_t=W_{t-1}-\eta\nabla\ell(W_{t-1};x_t),\end{equation}
where $\eta$ denotes the learning rate. In the basic naive version, self-supervised loss $\ell$ designed to restore corrupted input $\tilde{x}_t$. Although this straightforward reconstruction strategy proves beneficial in specific contexts, it exhibits constraints when modeling intricate relationships within input sequences, particularly for tasks demanding deeper contextual comprehension.
To overcome these constraints, we follow an advanced self-supervised task that exploits multiple views of the input data. We introduce learnable projection matrices $\theta_K$ and $\theta_V$ to map input data into different feature spaces. The training view $K=\theta_Kx_t$ extracts essential learning features while the label view $V=\theta_Vx_t$ serves as the reconstruction target:
\begin{equation}\ell(W;x_t)=\|f(K;W)-V\|^2,\end{equation}
this method enables the model to autonomously focus on the most critical features of the input data, significantly improving its ability to capture long-range dependencies and subtle relationships.

The final output feature $z_t$ is generated through the test view $Q=\theta_Qx_t$:
\begin{equation}z_t=f(Q;W_t),\end{equation}
where $f$ can be a linear model or multi-layer perceptron (we use a linear model in our implementation). The introduction of $\theta_Q$ endows the model with the ability to dynamically adjust feature attention during inference and automatically focus on the most discriminative features for different test samples, thereby greatly enhancing model adaptability to new data.
Finally, the output from the TTT layer and the result from the fourth branch are combined using the Hadamard product, followed by linear transformation and reshaping to meet the required dimensions for the Decoder’s subsequent layers.

\section{Experiments}

\begin{table*}[h!]
\centering
\caption{Configurations for each dataset. Patch Size refers to spatial dimensions of input patches (height, width) for 2D or (depth, height, width) for 3D datasets. "\# Stages" denotes the number of resolution states in the network hierarchy. "\# Pooling per axis" represents the number of downsampling operations along different spatial axes.}
\setlength{\tabcolsep}{10pt}
\begin{NiceTabular}{lcccc}
\toprule
\textbf{Configurations} & \textbf{Patch Size} & \textbf{Batch Size} & \textbf{\# Stages} & \textbf{\# Pooling per axis} \\
\midrule
ISIC2017 & (256, 256) & 8 & 4 & (6, 6) \\
Abdomen CT & (40, 224, 192) & 2 & 6 & (3, 3, 5) \\
3D Abdomen MR & (48, 160, 224) & 2 & 6 & (3, 3, 5) \\
Endoscopy & (384, 640) & 13 & 7 & (6, 6) \\
Microscopy & (512, 512) & 12 & 8 & (7, 7) \\
\bottomrule
\end{NiceTabular}
\label{tab:configurations}
\end{table*}
\begin{figure*}[!t]
\centerline{\includegraphics[width=\textwidth]{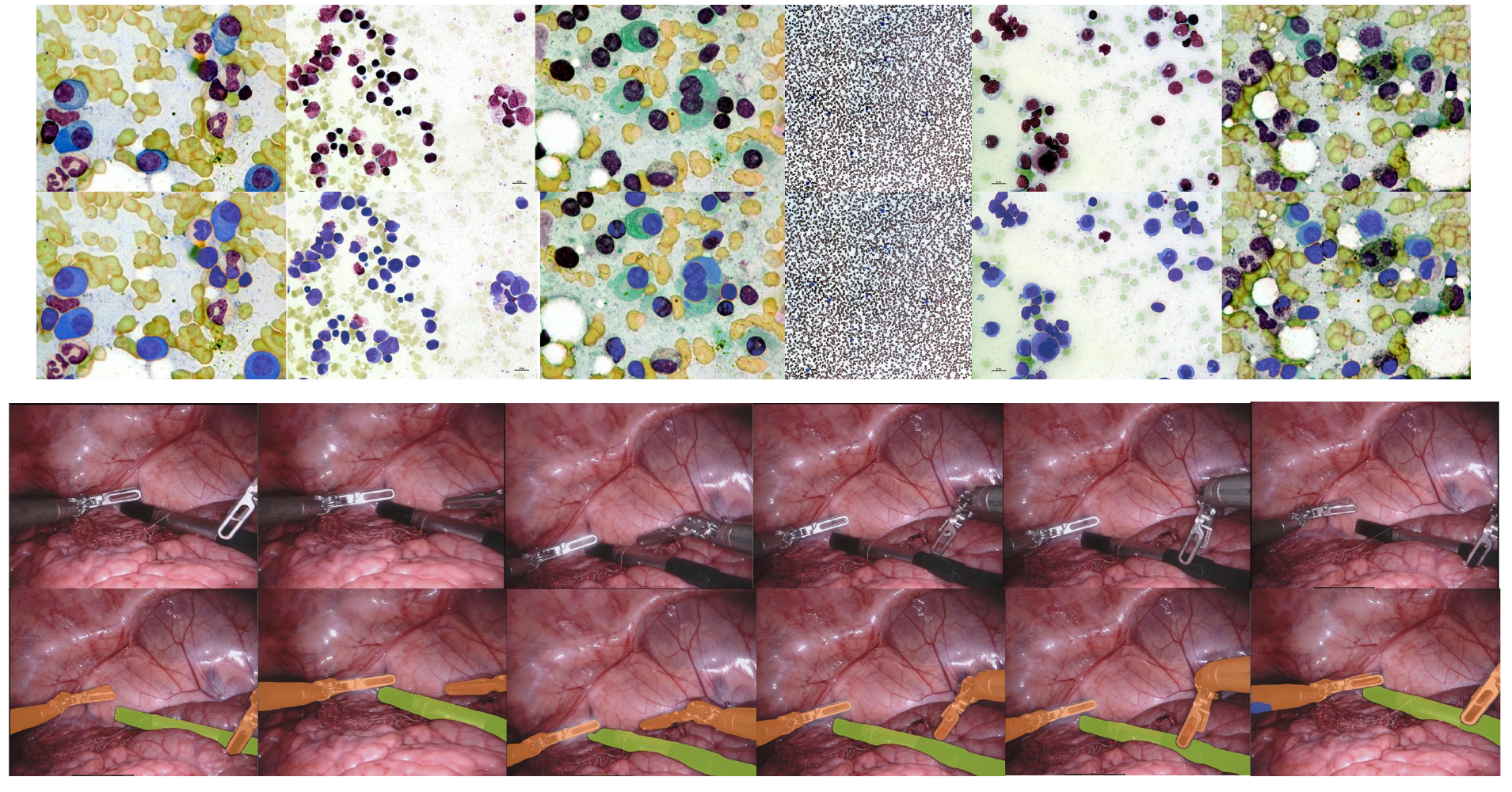}}
\caption{Visualized segmentation examples of cell segmentation in microscopy images (1st row), MedNext network + DSC block (2nd row), instruments segmentation in endoscopy images (3rd row), and U-Mamba network + DSC block (4th row).}
\label{Cell and instruments}
\end{figure*}
\subsection{Datasets}
To comprehensively assess the performance and scalability of the proposed DSC block, we conducted extensive experiments across five diverse medical image datasets encompassing various imaging modalities, dimensionalities, and segmentation objectives. Table \ref{tab:dataset information} summarizes the essential characteristics of these datasets.

\subsubsection{ISIC 2017}
The ISIC 2017~\cite{DBLP:journals/corr/abs-1710-05006} dataset is a comprehensive collection of 2,150 dermoscopic images of skin lesions, serving as a valuable resource for advancements in dermatology and computer-aided diagnosis. It encompasses a
broad spectrum of skin conditions, including both benign and malignant lesions, and is primarily designed for tasks related to melanoma detection and skin cancer classification. 
\subsubsection{Endoscopy images}
From the MICCAI 2017 EndoVis Challenge~\cite{DBLP:journals/corr/abs-1902-06426}, this dataset focuses on instrument segmentation within endoscopic images, featuring seven distinct instruments: the large needle driver, prograsp forceps, monopolar curved scissors, cadiere forceps, bipolar forceps, vessel sealer, and a drop-in ultrasound probe. The dataset is divided into 1,800 training frames and 1,200 testing frames.
\subsubsection{Microscopy images}
This dataset, from the NeurIPS 2022 Cell Segmentation Challenge~\cite{DBLP:journals/corr/abs-2308-05864}, is designed for cell segmentation in microscopy images, comprising 1,000 training images and 101 testing images. Following U-Mamba~\cite{DBLP:journals/corr/abs-2401-04722}, we treat this as a semantic segmentation task, focusing on cell boundaries and interiors rather than instance segmentation.
\subsubsection{Abdomen CT}
The Abdomen CT dataset (Ma et al. 2024b), part of the MICCAI 2022 FLARE challenge, includes the segmentation of 13 abdominal organs from 50 CT scans in both the training and testing sets. The segmented organs include the liver, spleen, pancreas, kidneys, stomach, gallbladder, esophagus, aorta, inferior vena cava, adrenal glands, and duodenum. 
\subsubsection{Abdomen MRI}
The Abdomen MR dataset (Ji et al. 2022), from the MICCAI 2022 AMOS Challenge, focuses on the seg-
mentation of the same 13 abdominal organs using MRI scans. It comprises 60 MRI scans for training and 50 for testing.
Additionally, we generate a 2D version of the dataset by converting the 3D abdominal MRI scans into 2D slices. This conversion allows us to evaluate DSC block within the common 2D segmentation framework, which is widely used in practice due
to its lower computational requirements. The conversion retains the same 13 organs, ensuring consistent evaluation across both 2D and 3D modalities.

\subsection{Implementation Details}
The specific configurations are comprehensively outlined in Table \ref{tab:configurations}. To balance computational efficiency with performance gains, we integrate the DSC block exclusively at the bottleneck layer (the deepest connection between encoder and decoder where feature dimensions are minimal but semantic information is richest) . This strategic placement minimizes parameter overhead (as shown in Table \ref{Flops}) while maximizing adaptive feature refinement at the most critical transmission point. This implementation was consistent across all six evaluated architectures: two CNN-based segmentation networks (nnU-Net~\cite{nnU-Net} and SegResNet~\cite{segresnet}), two Transformer-based networks (UNETR~\cite{Unetr} and SwinUNETR~\cite{SwinUnetr}), one hybrid CNN-Transformer network (MedNext~\cite{MedNexXt}), and one state-of-the-art Mamba-based network (U-Mamba~\cite{DBLP:journals/corr/abs-2401-04722}). 

\begin{table*}
\centering
\caption{Computational complexity comparison of baseline networks before and after integrating DSC block on Abdomen MRI dataset. The DSC block is integrated only at the bottleneck layer following the configuration in Section 4.2. The reported inference time is measured on a per-sample basis.}
\small
\setlength{\tabcolsep}{10pt}
\begin{tabular}{
    >{\centering\arraybackslash}m{3cm}
    c
    S[table-format=2.1]
    S[table-format=3.1]
    S[table-format=3.2]
}
\toprule
Model & DSC block & {Params (M)} & {GFLOPs} & {Inference Time (ms)} \\
\midrule
\multirow{2}{*}{nnU-Net}
 & $\times$      & 31.2 & 524.8 &  37.26 \\
 & $\checkmark$ & 32.6 & 525.3 &  41.30 \\
\midrule
\multirow{2}{*}{SegResNet}
 & $\times$      & 18.8 & 636.8 & 102.70 \\
 & $\checkmark$ & 19.2 & 645.4 & 290.04 \\
\midrule
\multirow{2}{*}{UNETR}
 & $\times$      & 93.0 & 194.5 &  39.46 \\
 & $\checkmark$ & 94.5 & 195.6 &  83.91 \\
\midrule
\multirow{2}{*}{SwinUNETR}
 & $\times$      & 62.1 & 830.7 & 165.44 \\
 & $\checkmark$ & 67.8 & 831.1 & 168.20 \\
\midrule
\multirow{2}{*}{MedNext}
 & $\times$      & 11.1 & 309.4 & 174.40 \\
 & $\checkmark$ & 12.2 & 313.1 & 216.22 \\
\midrule
\multirow{2}{*}{U-Mamba\_Bot}
 & $\times$      & 41.9 & 792.1 & 100.54 \\
 & $\checkmark$ & 43.3 & 792.3 & 109.31 \\
\bottomrule
\end{tabular}
\label{Flops}
\end{table*}

Training employed a combination of Dice loss and cross-entropy loss with SGD optimizer (initial learning rate: 1e-2) for 1000 epochs on a single A100 GPU. Network blocks were automatically configured using nnU-Net's self-configuration capability. Performance evaluation followed the protocol established in U-Mamba~\cite{DBLP:journals/corr/abs-2401-04722}: Dice Similarity Coefficient (Dice score) and Normalized Surface Distance (NSD) for ISIC2017 skin lesion segmentation, abdominal organ segmentation, and endoscopic instrument segmentation. F1 score was employed for microscopy cell segmentation tasks. The TTT module employs several key design choices to ensure both effectiveness and stability. Following the mini-batch TTT strategy~\cite{2024Learning}, the number of gradient update steps during inference is determined by $T/b$, where $T$ is the the feature sequence length and b is the mini-batch size. In our implementation, we use b=16 to balance computational efficiency and adaptation quality. The hidden state dimensions are controlled by a reduction ratio r=16, where the intermediate dimension is computed as $d=max(C/r,32)$ with $C$ being the number of input channels. The projection matrices $\theta_K$, $\theta_V$, $\theta_Q$ are implemented as $1\times1\times1$ convolutional layers that map input features from dimension C to the reduced dimension d.

\subsection{Quantitative and Qualitative Segmentation Results}
\subsubsection{2D Dataset Results}
\begin{table*}[h!]
\centering
\caption{Results summary of 2D segmentation tasks. In the DSC block column, ``-'' indicates methods unrelated to DSC block integration, ``$\times$'' indicates baseline methods without DSC block, and ``$\checkmark$'' indicates methods with DSC block integrated. ``*'' and ``$\dagger$'' means the experimental results are from existing works, where ``*'' are from~\cite{DBLP:journals/corr/abs-2401-04722}, ``$\dagger$'' are from~\cite{zhou2024tttunetenhancingunettesttime}.}
\setlength{\tabcolsep}{6.5pt}
\begin{NiceTabular}{c|cc|cc|cc|c}
    \toprule
    \multirow{2}[4]{*}{Methods} & \multicolumn{2}{c|}{\multirow{2}[4]{*}{DSC block}} & \multicolumn{2}{c|}{Instruments in Endoscopy} & \multicolumn{2}{c}{ISIC 2017} & \multicolumn{1}{c}{Cells in Microscopy} \\
    \cmidrule{4-5} \cmidrule{6-7} \cmidrule{8-8}
          & \multicolumn{2}{c|}{} & \multicolumn{1}{c}{Dice} & \multicolumn{1}{c|}{NSD} & \multicolumn{1}{c}{Dice} & \multicolumn{1}{c|}{NSD} & \multicolumn{1}{c}{F1}  \\
    \midrule
    Med-TTT & \multicolumn{2}{c|}{$-$} & 0.5011±0.3168\hspace{0.5em} & 0.5319±0.3033\hspace{0.5em} & 0.8149±0.3105 & 0.8642±0.2974 & 0.4018±0.2239\hspace{0.5em} \\
    \midrule
    TTT-UNet\_Bot & \multicolumn{2}{c|}{$-$} & 0.6643±0.3018$^{\dagger}$ & 0.6799±0.3056$^{\dagger}$ & 0.8878±0.3216 & 0.9103±0.3164 & 0.5818±0.2410$^{\dagger}$ \\
    \midrule
    \multirow{2}[2]{*}{nnU-Net} & \multicolumn{2}{c|}{$\times$} & 0.6264±0.3024$^*$ & 0.6412±0.3074$^*$ & 0.8825±0.2739 & 0.9120±0.3110 & 0.5383±0.2657$^*$ \\
        & \multicolumn{2}{c|}{$\checkmark$} & \textbf{0.6718±0.3221}\hspace{0.5em} & \textbf{0.6865±0.3182}\hspace{0.5em} & \textbf{0.8846±0.2634} & \textbf{0.9210±0.3354} & \textbf{0.5460±0.2653}\hspace{0.5em} \\
    \midrule
    \multirow{2}[2]{*}{SegResNet} & \multicolumn{2}{c|}{$\times$} & 0.5820±0.3268$^*$ & 0.5968±0.3303$^*$ & 0.8763±0.3521 & 0.9070±0.3619 & 0.5411±0.2633$^*$ \\
        & \multicolumn{2}{c|}{$\checkmark$} & \textbf{0.6620±0.3029}\hspace{0.5em} & \textbf{0.6782±0.3410}\hspace{0.5em} & \textbf{0.8796±0.3211} & \textbf{0.9108±0.3117} & \textbf{0.5779±0.2748}\hspace{0.5em} \\
    \midrule
    \multirow{2}[2]{*}{UNETR} & \multicolumn{2}{c|}{$\times$} & 0.5017±0.3201$^*$ & 0.5168±0.3235$^*$ & 0.8499±0.3328 & 0.8850±0.3198 & 0.4357±0.2572$^*$ \\
        & \multicolumn{2}{c|}{$\checkmark$} & \textbf{0.5137±0.3320}\hspace{0.5em} & \textbf{0.5275±0.3302}\hspace{0.5em} & \textbf{0.8580±0.3019} & \textbf{0.8930±0.3164} & \textbf{0.4856±0.2637}\hspace{0.5em} \\
    \midrule
    \multirow{2}[2]{*}{SwinUNETR} 
        & \multicolumn{2}{c|}{$\times$} & 0.5528±0.3089$^*$ & 0.5683±0.3123$^*$ & 0.8803±0.2482 & 0.9145±0.3216 & 0.3967±0.2621$^*$ \\
        & \multicolumn{2}{c|}{$\checkmark$} & \textbf{0.6064±0.3408}\hspace{0.5em} & \textbf{0.6226±0.3356}\hspace{0.5em} & \textbf{0.8820±0.3190} & \textbf{0.9156±0.2682} & \textbf{0.4807±0.2375}\hspace{0.5em} \\
    \midrule
    \multirow{2}[2]{*}{MedNext} 
        & \multicolumn{2}{c|}{$\times$} & 0.6055±0.3476\hspace{0.5em} & 0.6211±0.3329\hspace{0.5em} & 0.8871±0.2398 & 0.9212±0.3182 & 0.5613±0.2573\hspace{0.5em} \\
        & \multicolumn{2}{c|}{$\checkmark$} & \textbf{0.6104±0.3528}\hspace{0.5em} & \textbf{0.6263±0.3427}\hspace{0.5em} & \textbf{0.8901±0.3821} & \textbf{0.9223±0.3742} & \textbf{0.5746±0.2469}\hspace{0.5em} \\
    \midrule
    \multirow{2}[2]{*}{U-Mamba\_Bot} 
        & \multicolumn{2}{c|}{$\times$} & 0.6540±0.3008$^*$ & 0.6692±0.3050$^*$ & 0.8889±0.3471 & 0.9173±0.3284 & 0.5389±0.2817$^*$ \\
        & \multicolumn{2}{c|}{$\checkmark$} & \textbf{0.6733±0.3217}\hspace{0.5em} & \textbf{0.6887±0.3054}\hspace{0.5em} & \textbf{0.8892±0.3328} & \textbf{0.9182±0.3165} & \textbf{0.6101±0.2465}\hspace{0.5em} \\
    \bottomrule
    \end{NiceTabular}
\label{tab:2D}
\end{table*}

\begin{figure*}[!t]
\centerline{\includegraphics[width=\textwidth]{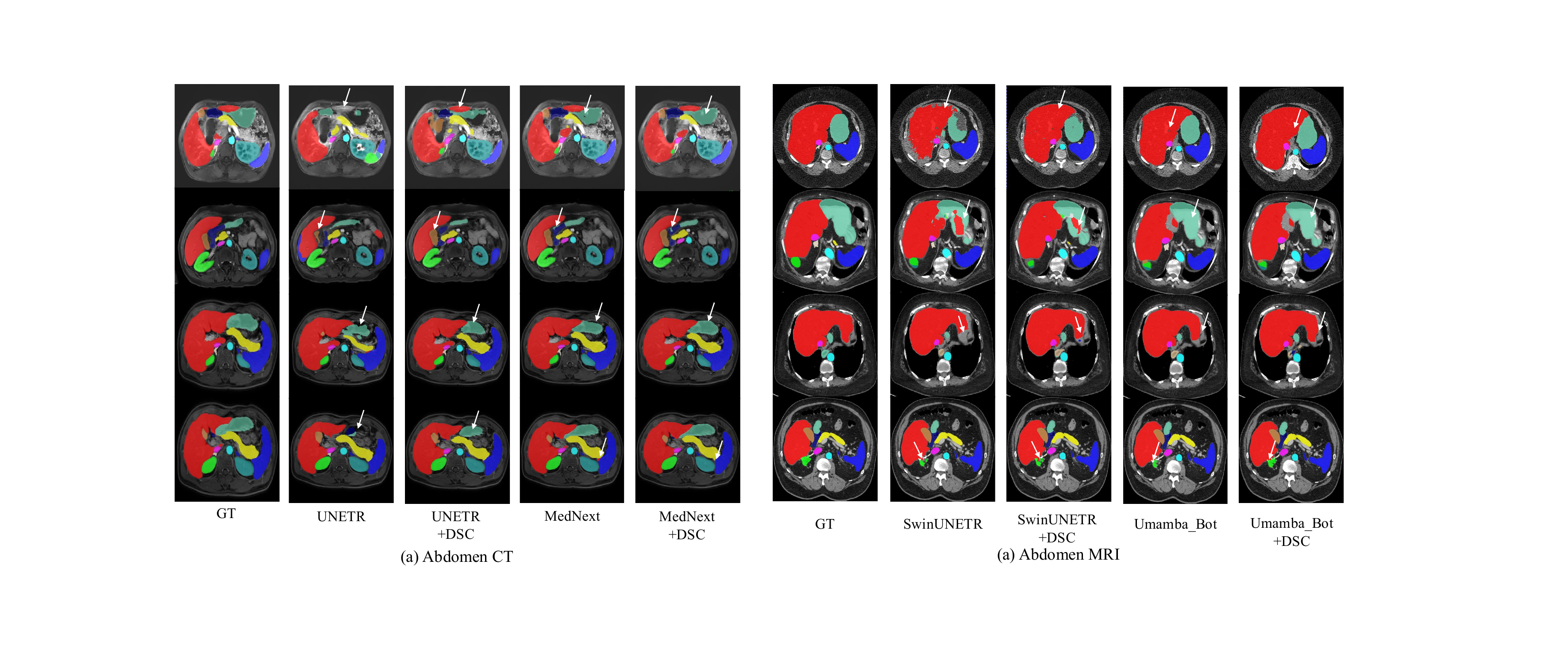}}
\caption{Visualized segmentation examples of abdominal organ segmentation in CT(a) and MRI(b).}
\label{CT and MRI}
\end{figure*}

\begin{table*}[h!]
\centering
\caption{Results summary of 3D segmentation tasks. ``*'' and ``$\dagger$'' means the experimental results are from existing works, where ``*'' are from~\cite{DBLP:journals/corr/abs-2401-04722}, ``$\dagger$'' are from~\cite{zhou2024tttunetenhancingunettesttime}.}
\setlength{\tabcolsep}{13pt}
\begin{NiceTabular}{c|cc|cc|cc}
    \toprule
    \multirow{2}[4]{*}{Methods} & \multicolumn{2}{c|}{\multirow{2}[4]{*}{DSC block}} & \multicolumn{2}{c|}{Organs in Abdomen CT} & \multicolumn{2}{c}{Organs in Abdomen MRI} \\
    \cmidrule{4-5} \cmidrule{6-7}
          & \multicolumn{2}{c|}{} & \multicolumn{1}{c}{Dice} & \multicolumn{1}{c|}{NSD} & \multicolumn{1}{c}{Dice} & \multicolumn{1}{c}{NSD} \\
    \midrule
    TTT-UNet\_Bot & \multicolumn{2}{c|}{$-$} &0.8709±0.1011$^{\dagger}$  &0.8995±0.0721$^{\dagger}$   &0.8677±0.0482$^{\dagger}$   &0.9247±0.0631$^{\dagger}$  \\
    \midrule
    \multirow{2}[2]{*}{nnU-Net} & \multicolumn{2}{c|}{$\times$} & 0.8615±0.0790$^*$ & 0.8972±0.0824$^*$ & 0.8309±0.0769$^*$ & 0.8996±0.0729$^*$ \\
        & \multicolumn{2}{c|}{$\checkmark$} & \textbf{0.8718±0.1177}\hspace{0.5em} & \textbf{0.9093±0.0317}\hspace{0.5em} & \textbf{0.8461±0.1062}\hspace{0.5em} & \textbf{0.9133±0.0401}\hspace{0.5em} \\
    \midrule
    \multirow{2}[2]{*}{SegResNet} & \multicolumn{2}{c|}{$\times$} & 0.7927±0.1162$^*$ & 0.8257±0.1194$^*$ & \textbf{0.8146±0.0959}$^*$ & \textbf{0.8841±0.0917}$^*$ \\
        & \multicolumn{2}{c|}{$\checkmark$} & \textbf{0.8054±0.1700}\hspace{0.5em} & \textbf{0.8439±0.0147}\hspace{0.5em} & 0.8136±0.0963\hspace{0.5em} & 0.8795±0.0928\hspace{0.5em} \\
    \midrule
    \multirow{2}[2]{*}{UNETR} & \multicolumn{2}{c|}{$\times$} & 0.6824±0.1506$^*$ & 0.7004±0.1577$^*$ & 0.6867±0.1488$^*$ & 0.7440±0.1627$^*$ \\
        & \multicolumn{2}{c|}{$\checkmark$} & \textbf{0.6972±0.2410}\hspace{0.5em} & \textbf{0.7265±0.0865}\hspace{0.5em} & \textbf{0.6971±0.0591}\hspace{0.5em} & \textbf{0.7576±0.0453}\hspace{0.5em} \\
    \midrule
    \multirow{2}[2]{*}{SwinUNETR} 
        & \multicolumn{2}{c|}{$\times$} & 0.7594±0.1095$^*$ & 0.7663±0.1190$^*$ & \textbf{0.7565±0.1394}$^*$ & \textbf{0.8218±0.1409}$^*$ \\
        & \multicolumn{2}{c|}{$\checkmark$} & \textbf{0.7847±0.2005}\hspace{0.5em} & \textbf{0.8179±0.0411}\hspace{0.5em} & 0.7469±0.0655\hspace{0.5em} & 0.8091±0.0467\hspace{0.5em} \\
    \midrule
    \multirow{2}[2]{*}{MedNext} 
        & \multicolumn{2}{c|}{$\times$} & 0.8427±0.1137\hspace{0.5em} & 0.8706±0.1538\hspace{0.5em} & 0.8270±0.0625\hspace{0.5em} & 0.8951±0.0428\hspace{0.5em} \\
        & \multicolumn{2}{c|}{$\checkmark$} & \textbf{0.8642±0.0786}\hspace{0.5em} & \textbf{0.8920±0.0412}\hspace{0.5em} & \textbf{0.8421±0.1098}\hspace{0.5em} & \textbf{0.9065±0.0667}\hspace{0.5em} \\
    \midrule
    \multirow{2}[2]{*}{U-Mamba\_Bot} 
        & \multicolumn{2}{c|}{$\times$} & 0.8683±0.0808$^*$ & \textbf{0.9049±0.0821}$^*$ & 0.8453±0.0673$^*$ & 0.9121±0.0634$^*$ \\
        & \multicolumn{2}{c|}{$\checkmark$} & \textbf{0.8712±0.0973}\hspace{0.5em} & 0.9037±0.0321\hspace{0.5em} & \textbf{0.8535±0.0597}\hspace{0.5em} & \textbf{0.9172±0.0340}\hspace{0.5em} \\
    \bottomrule
    \end{NiceTabular}
\label{tab:3D}
\end{table*}

For 2D tasks, we report Dice score, NSD, and F1 scores as percentages using cross-validation, following the protocol of the backbone networks. Table \ref{tab:2D} presents the performance of DSC block integration across multiple 2D segmentation tasks.
TTT-UNet\_Bot and Med-TTT are recent works that also apply TTT to medical image segmentation tasks. Med-TTT only provides 2D results, as its implementation supports 2D data processing and does not provide 3D functionality in their released code.
The DSC block demonstrates consistent improvements across all evaluated architectures. For endoscopy instrument segmentation, U-Mamba\_Bot with DSC achieves the highest Dice score of 0.6733±0.3217, substantially improving over baseline (0.6540±0.3008). nnU-Net with DSC reaches 0.6718±0.3221 compared to 0.6264±0.3024 without DSC. SegResNet shows remarkable enhancement from 0.5820±0.3268 to 0.6620±0.3029. The consistent performance gains across CNN-based architectures demonstrate the broad compatibility of our approach with traditional convolutional frameworks.
Transformer-based architectures also benefit significantly from DSC integration. UNETR advances from 0.5017±0.3201 to 0.5137±0.3320, while SwinUNETR shows improvement from 0.5528±0.3089 to 0.6064±0.3408. These enhancements are particularly noteworthy given that Transformer-based methods already incorporate sophisticated global attention mechanisms, suggesting that our dynamic skip connection approach provides complementary benefits to existing self-attention computations. 
For ISIC2017 skin lesion segmentation, SwinUNETR with DSC achieves 0.8820±0.3190, while MedNext with DSC reaches 0.8901±0.3821. The most pronounced improvements occur in cell segmentation, where U-Mamba\_Bot with DSC achieves F1 score of 0.6101±0.2465, substantially outperforming baseline (0.5389±0.2817). nnU-Net demonstrates improvement from 0.5383±0.2657 to 0.5460±0.2653, while SegResNet shows enhancement from 0.5411±0.2633 to 0.5779±0.2748. The substantial improvement in UNETR performance, advancing from 0.4357±0.2572 to 0.4856±0.2637, indicates that dynamic adaptation is particularly beneficial for architectures that may struggle with fine-grained cellular feature discrimination.

Fig. \ref{Cell and instruments} provides comprehensive qualitative analysis across different 2D segmentation scenarios. For microscopic cell segmentation (top row), the visualizations reveal diverse cellular morphologies with varying sizes, densities, and clustering patterns. The enhanced segmentation capability of DSC-enabled networks is particularly evident in regions with overlapping cells and ambiguous boundaries. In the endoscopic instrument segmentation examples (bottom rows), the DSC block demonstrates superior performance in delineating surgical instruments against complex tissue backgrounds. The adaptive kernel selection mechanism effectively handles the varying scales of instruments, from fine-tipped graspers to broader surgical paddles, while the TTT module enables real-time adaptation to different lighting conditions and tissue appearances commonly encountered in endoscopic procedures.
These results demonstrate that the DSC block provides consistent performance gains across diverse 2D segmentation tasks, with particularly strong improvements in challenging scenarios involving complex morphological structures.

\subsubsection{3D Dataset Results}
For 3D tasks, we report Dice score and NSD scores using cross-validation, following the protocol of the backbone networks. Table \ref{tab:3D} presents the performance of DSC block integration on 3D organ segmentation across abdominal CT and MRI datasets.
The DSC block shows performance improvements across most evaluated methods. For abdominal CT segmentation, nnU-Net with DSC achieves 0.8718±0.1177 compared to baseline 0.8615±0.0790. U-Mamba\_Bot with DSC reaches 0.8712±0.0973, while baseline achieves 0.8683±0.0808. MedNext with DSC achieves 0.8642±0.0786 compared to baseline 0.8427±0.1137. SegResNet demonstrates significant enhancement from 0.7927±0.\linebreak1162 to 0.8054±0.1700, while UNETR shows improvement from 0.6824±0.1506 to 0.6972±0.2410. SwinUNETR advances from 0.7594±0.1095 to 0.7847±0.2005, and these consistent improvements across all architectures indicate that the DSC block provides universal benefits across different architectural paradigms in volumetric segmentation scenarios.
For abdominal MRI segmentation, U-Mamba\_Bot with DSC achieves 0.8535±0.0597 compared to baseline 0.8453±0.0673. nnU-Net with DSC reaches 0.8461±0.1062 compared to baseline 0.8309±0.0769. MedNext demonstrates improvement from 0.8270±0.0625 to 0.8421±0.1098, while UNETR advances from 0.6867±0.1488 to 0.6971±0.0591. Although SwinUNETR shows a slight decrease from 0.7565±\linebreak0.1394 to 0.7469±0.0655, the DSC block maintains stable and competitive performance across different MRI contrast patterns and imaging protocols.
Fig. \ref{CT and MRI} illustrates qualitative segmentation results on abdominal CT and MRI data. The systematic comparison presents ground truth annotations alongside baseline network outputs and their corresponding DSC-enhanced versions across four representative architectures: UNETR, MedNext, SwinUNETR, and U-Mamba. The visual evidence clearly demonstrates consistent improvements when DSC blocks are integrated into existing frameworks. The CT segmentation results reveal that DSC integration leads to more accurate organ boundary delineation and improved anatomical consistency. 

Across most evaluated architectures, the DSC-enhanced versions produce segmentations that more closely approximate the ground truth annotations, with notable improvements in maintaining proper organ connectivity and reducing segmentation artifacts. The MRI segmentation comparisons demonstrate the performance of DSC across different imaging modalities, with enhanced networks generally delivering improved segmentation quality and better handling of the complex tissue contrast patterns characteristic of MRI imaging. The results indicate that DSC integration addresses common segmentation challenges, including boundary ambiguities and intensity variations. The clinical significance of these improvements extends beyond quantitative metrics, as enhanced segmentation accuracy supports more reliable automated analysis in diagnostic workflows. Through comprehensive evaluation spanning multiple architectures and imaging technologies, the DSC block establishes itself as an effective enhancement strategy that maintains performance across diverse medical imaging scenarios.

\subsection{Ablation Study}
\subsubsection{Impact of Skip-Connection Enhancement Strategies}
This experiment was conducted on the ISIC2017 dataset to evaluate the effectiveness of our DSC block against existing skip connection enhancement methods. Table \ref{strategies} illustrates the performance when different skip connection enhancement strategies are applied. Our UNet+DSC achieves superior performance with mIoU of 0.7980 and Dice score of 0.8876, outperforming both direct modification methods including U-Net++ (mIoU: 0.7334, Dice: 0.8461) and UNet3+ (mIoU: 0.7805, Dice: 0.8767), as well as indirect enhancement approaches including TransUNet (mIoU: 0.7834, Dice: 0.8665) and HiFormer (mIoU: 0.7695, Dice: 0.8524). Notably, the improvements are consistent across both evaluation metrics. The results demonstrate that our DSC block provides more effective skip connection enhancement compared to existing static connection architectures and indirect encoder-based improvements.
\begin{table}[h!]
\centering
\setlength{\tabcolsep}{10pt}
\caption{Results summary of 2D skin lesion segmentation on ISIC2017 dataset.}
\begin{tabular}{lSSS}
\toprule
\textbf{Methods} & \textbf{mIoU} & \textbf{Dice} &\textbf{NSD} \\
\midrule
U-Net~\cite{2015U} & 0.7102 & 0.8355 & 0.5298\\
U-Net++~\cite{zhou2018unetplusplus} & 0.7334 & 0.8461 & 0.6265\\
SwinUNet~\cite{SwinUnet} & 0.6693 & 0.7782 & 0.5778\\
UNet3+~\cite{2020arXiv200408790H} & 0.7805 & 0.8767 & 0.6283 \\
TransUNet~\cite{transunet} & 0.7834 & 0.8665 & 0.6124 \\
Hiformer~\cite{DBLP:conf/wacv/HeidariKKAACM23} & 0.7695 & 0.8524 & 0.6031\\
FuseUNet~\cite{he2025fuseunetmultiscalefeaturefusion} & \textbf{0.8058} & 0.8773 & 0.6420  \\
\midrule
U-Net + DSC & 0.7980 & \textbf{0.8876} & \textbf{0.6423 }\\
\bottomrule
\end{tabular}
\label{strategies}
\end{table}

\subsubsection{Impact of DSC block Placement Strategy}
To evaluate the performance-efficiency trade-off of different DSC block integration strategies, we conducted an ablation study using UNETR as the baseline architecture on the ISIC 2017 dataset. We compared three configurations: (1) Baseline (no DSC), (2) Bottleneck-only (DSC integrated solely at the deepest skip connection), and (3) All levels (DSC integrated at every skip connection level). This allows us to assess the impact of placement on parameters, computational cost (GFLOPs), inference time, and segmentation metrics (Dice and NSD). Note that while the DSC block is designed as a general module applicable to any skip connection level, multi-level integration, though feasible and yielding slight performance gains, incurs significantly higher computational overhead due to repeated TTT updates across layers. This validates our strategic choice of bottleneck-only placement to balance efficacy and efficiency, avoiding misleading readers about the "dynamic skip connections" being limited to a single layer—multi-level application is possible but not recommended for practical deployment where inference time is critical.
As shown in Table \ref{all-levels}, the Bottleneck-only configuration achieves an optimal balance, with minimal increases in parameters (87.9M vs. baseline 87.7M) and GFLOPs (26.5 vs. 26.4), and a moderate rise in inference time (47.19 ms vs. 17.41 ms), while improving Dice to 0.8580±0.3019 (vs. baseline 0.8499±0.3328) and NSD to 0.8930±0.3164 (vs. baseline 0.8850±0.3198). The All levels configuration further enhances accuracy slightly, with Dice at 0.8625±0.2950 and NSD at 0.8975±0.3100, but escalates parameters to 88.1M, GFLOPs to 27.5, and inference time dramatically to 641.55 ms. This performance uplift does not justify the substantial overhead, confirming that bottleneck-only integration is preferable for real-time medical imaging applications, though multi-level placement remains viable for scenarios prioritizing maximal accuracy over efficiency.

\subsubsection{Impact of Kernel Size Configuration}
This experiment was conducted on the ISIC2017 dataset to examine the effect of different kernel combinations within the DMSK module. Table \ref{tab:7} shows that combining small and large kernels achieves optimal performance (mIoU: 0.7980, Dice: 0.8876), significantly outperforming small kernels only (mIoU: 0.7629, Dice: 0.8638) or large kernels only (mIoU: 0.7729, Dice: 0.8695). The results indicate that small kernels excel at capturing fine-grained local details and precise boundary information, while large kernels provide superior global contextual understanding and long-range spatial dependencies. The performance gap between single-scale configurations and the combined approach highlights the importance of multi-scale feature integration in medical image segmentation tasks. This demonstrates that multi-scale kernel integration effectively captures complementary information, with small kernels providing fine-grained details and large kernels offering global context.
\subsubsection{Impact of Individual Module Components}
This experiment evaluates individual contributions of DMSK and TTT modules on Organs in Abdomen MRI (3D) and Instruments in Endoscopy (2D) datasets. Table \ref{Ablation on UNETR} demonstrates each component's impact on segmentation performance. On the Abdomen MRI dataset, baseline UNETR achieves 0.6867±0.1488 (Dice) and 0.7440±0.1627 (NSD). Individual modules provide incremental improvements: DMSK alone achieves 0.6943±0.1408 Dice and 0.7534±0.1603 NSD, while TTT alone reaches 0.6965±0.153\linebreak8 Dice and 0.7546±0.1548 NSD, with DMSK addressing intra-feature constraints through adaptive multi-scale feature extraction and TTT resolving inter-feature constraints through dynamic weight adaptation. The combined implementation yields optimal performance at 0.6971±0.1533 Dice and 0.7576±0.1653 NSD. On the Endoscopy dataset, individual modules show progressive gains over baseline (0.5017±0.3201 Dice, 0.5168±0.3235 NSD), with DMSK achieving 0.5038±0.3148 Dice and 0.5237±0.2365 NSD, and TTT reaching 0.5042±0.2785 Dice and 0.5259±0.2876 NSD. The combination achieves highest performance at 0.5137±0.3320 Dice and 0.5275±0.3302 NSD. These results indicate that both modules contribute complementary enhancements, with their integration consistently outperforming individual components across imaging modalities.

\begin{table*}
\centering
\caption{Performance-efficiency trade-off of DSC block placement strategies on ISIC 2017. UNETR is used as the baseline architecture. Bottleneck-only integration demonstrates optimal balance between accuracy gains and computational overhead.}
\setlength{\tabcolsep}{15pt}
\small 
\begin{tabular}{
    l
    S[table-format=2.1]
    S[table-format=3.1]
    S[table-format=3.2]
    c
    c
    }
\hline
Configuration & {Params(M)} & {GFLOPs} & {Inference Time (ms)} &{Dice} & {NSD}\\
\hline
Baseline       & 87.7  & 26.4 &  17.41 & 0.8499±0.3328 &0.8850±0.3198 \\
\hline
Bottleneck-only     & 87.9  & 26.5   & 47.19  & 0.8580±0.3019 &0.8930±0.3164 \\
\hline
All levels & 88.1  & 27.5  & 641.55   & 0.8625±0.2950 &0.8975±0.3100 \\
\hline
\end{tabular}
\label{all-levels}
\end{table*}

\begin{table}[!htbp]
\centering
\setlength{\tabcolsep}{3pt}
\caption{Kernel size ablation study on ISIC2017 dataset with UNet + DSC.}
\begin{tabular}{lcSSS}
\toprule
\textbf{Configuration} & \textbf{Kernel Strategy} & \textbf{mIoU} & \textbf{Dice} & \textbf{NSD}  \\
\midrule
\multirow{3}{*}{UNet + DSC} 
    & Small \& Large Kernels & 0.7980 & 0.8876 & 0.6423\\
    & Small Kernels Only & 0.7629 & 0.8638 & 0.6418\\
    & Large Kernels Only & 0.7729 & 0.8695 & 0.6420\\
\bottomrule
\end{tabular}
\label{tab:7}
\end{table}

\begin{table}[!htbp]
\centering
\setlength{\tabcolsep}{2.5pt}
\caption{Component ablation study on UNETR with DSC block.}
\begin{tabular}{lcccc}
\toprule
\textbf{Dataset} & \textbf{DMSK} & \textbf{TTT} & \textbf{Dice} & \textbf{NSD} \\
\midrule
\multirow{4}{*}{\makecell{Organs in \\ Abdomen MRI}}
    & $\times$ & $\times$ & 0.6867±0.1488 & 0.7440±0.1627 \\
    & $\checkmark$ & $\times$ & 0.6943±0.1408 & 0.7534±0.1603 \\
    & $\times$ & $\checkmark$ & 0.6965±0.1538 & 0.7546±0.1548 \\
    & $\checkmark$ & $\checkmark$ & \textbf{0.6971±0.1533} & \textbf{0.7576±0.1653} \\
\midrule
\multirow{4}{*}{\makecell{Instruments in \\ Endoscopy}}
    & $\times$ & $\times$ & 0.5017±0.3201 & 0.5168±0.3235 \\
    & $\checkmark$ & $\times$ & 0.5038±0.3148 & 0.5237±0.2365 \\
    & $\times$ & $\checkmark$ & 0.5042±0.2785 & 0.5259±0.2876 \\
    & $\checkmark$ & $\checkmark$ & \textbf{0.5137±0.3320} & \textbf{0.5275±0.3302} \\
\bottomrule
\end{tabular}
\label{Ablation on UNETR}
\end{table}

\begin{table}
\centering
\setlength{\tabcolsep}{5.5pt}
\caption{Ablation study comparing cascade and parallel multi-scale fusion strategies in DMSK on the ISIC 2017 dataset.}
\begin{tabular}{lcSSS}
\toprule
\textbf{Method}  & \textbf{Kernel Strategy} &\textbf{mIoU} & \textbf{Dice} & \textbf{NSD}  \\
\midrule
\multirow{2}{*}{UNet + DSC} 
    & Parallel & 0.7830 & 0.8827 & 0.6257\\
    & Cascade (ours) & 0.7980 & 0.8876 & 0.6423\\
\bottomrule
\end{tabular}
\label{tab:cascade}
\end{table}

\begin{figure}
\centering
\includegraphics[width=1\columnwidth]{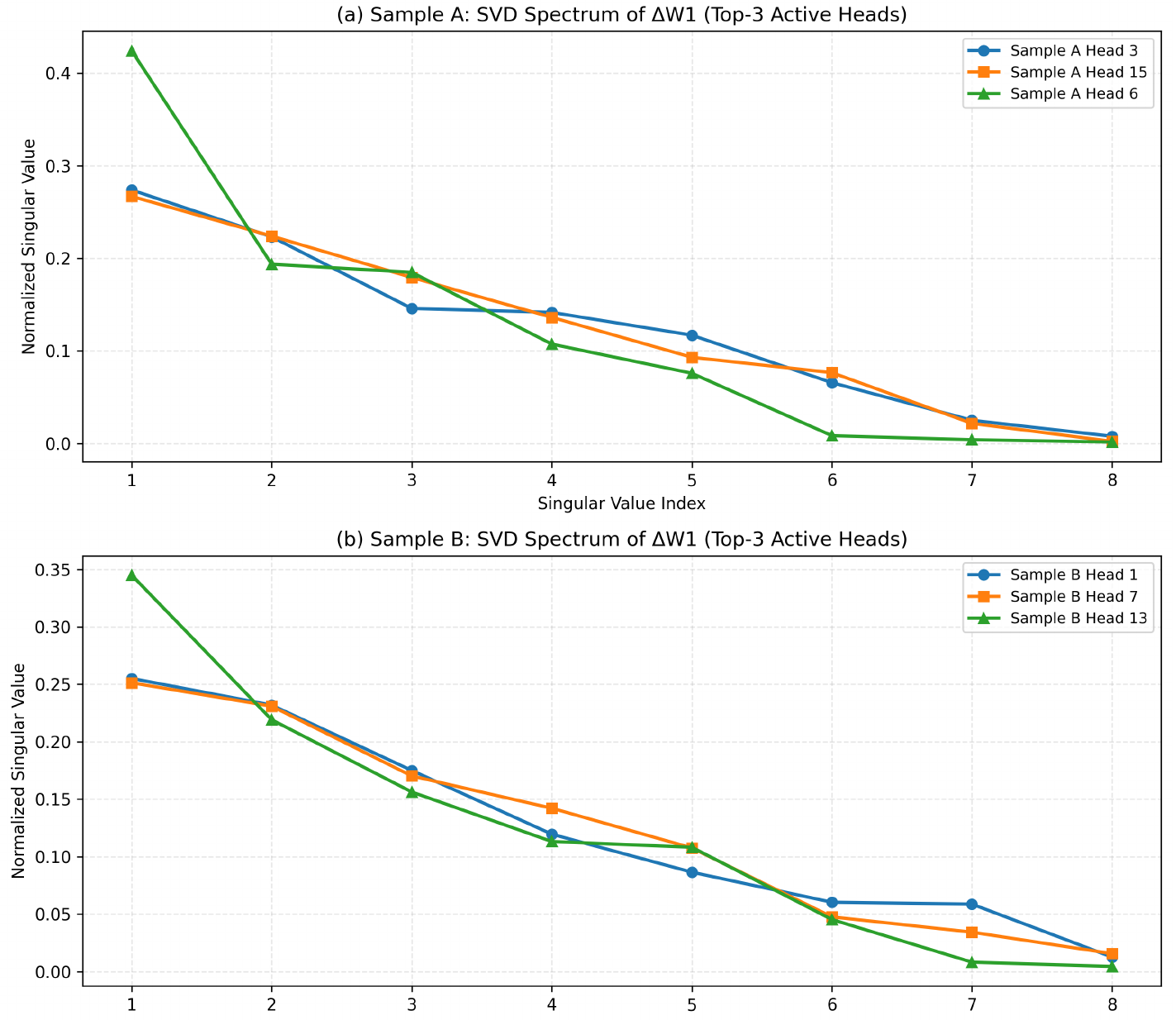}
\caption{Singular value spectrum of weight updates $\Delta \mathrm{W}_1$ for (a) a challenging sample and (b) an easier sample. Larger singular values in the challenging case demonstrate input-dependent adaptation, while rapid decay in both cases indicates targeted, low-rank updates.}
\label{weight}
\end{figure}

\subsubsection{Impact of Multi-Scale Fusion Strategy}
To justify the sequential (cascade) application of kernels in the DMSK module, we conducted an ablation study comparing it with a parallel multi-branch variant, where the small-scale kernel $k_s$ and large-scale kernel $k_b$ operate independently on the input features before concatenation. This parallel design follows common multi-scale fusion practices in the literature, but differs from our recursive aggregation in Equation (8), which enables hierarchical refinement by processing local details first and then integrating broader context.
The experiment was performed on the ISIC 2017 dataset using UNet as the baseline backbone with DSC integration. As reported in Table \ref{tab:cascade}, the cascade strategy outperforms the parallel variant across all metrics. These gains demonstrate that the sequential fusion promotes more effective multi-scale interaction, allowing fine-grained local features from $k_s$ to inform the global contextual extraction in $k_b$, which is particularly beneficial for handling varying lesion scales and complex boundaries in medical images. In contrast, the parallel approach may introduce redundancy and less adaptive refinement, leading to suboptimal feature integration.

\section{Conclusion}
Our DSC block transforms static skip connections into adaptive mechanisms that respond to unique image characteristics during inference. As a plug-and-play component, it seamlessly incorporates into existing U-like network structures without modifications. The TTT is applied to skip connection, allowing targeted feature adaptation, as illustrated in Fig. \ref{weight}. Our work moves beyond empirical network design by identifying and addressing fundamental limitations in skip connection mechanisms. The DSC block fundamentally enhances cross-layer connectivity through adaptive mechanisms, demonstrating superior feature representation capabilities. This approach is particularly valuable when handling the substantial heterogeneity present in medical imaging data. However, a common limitation across TTT-enhanced methods is the inherent computational overhead introduced by dynamic adaptation mechanisms. The TTT module requires additional processing during inference, potentially increasing latency in time-sensitive clinical applications. This computational burden represents a persistent trade-off between adaptivity and efficiency in existing TTT-based approaches. Future research should prioritize developing computationally efficient TTT implementations and lightweight dynamic adaptation strategies. Such developments are crucial for translating adaptive mechanisms into real-world clinical applications where both accuracy and computational efficiency are paramount.

\section*{Acknowledgements}
This work was supported by the Major Program of the National Natural Science Foundation of China under Grant No. 62495064 and the Youth Program of the National Natural Science Foundation of China under Grant No. 62206189.

\bibliographystyle{unsrt}
\bibliography{ref}



\end{document}